\documentclass[letterpaper, 10 pt, conference]{ieeeconf}

\IEEEoverridecommandlockouts                              % This command is only needed if 
\usepackage[T1]{fontenc}

\usepackage{amsfonts}	% Additional math fonts
\usepackage{amsmath}	% Math symbols
\usepackage{amssymb}    % Extended math symbols
\usepackage{siunitx}
\usepackage{pifont}   % check mark
\usepackage{dsfont}   % indicator function

\usepackage{booktabs}
\usepackage{makecell}  % Line breaks in table cells
\usepackage[flushleft]{threeparttable}  % For notes below a table
\usepackage{multirow}
\usepackage{rotating}

\usepackage{xspace}    % for correct spacing after defined commands without arguments, e.g., \net
\usepackage{xcolor}    % for colored text
\usepackage{colortbl}
\let\labelindent\relax
\usepackage[inline]{enumitem} % To change label for enumerates
\usepackage{textcomp}  % prevent warnings of gensymb package
\usepackage{gensymb}   % \degree
\usepackage{graphicx} % insert PDF as figure
\usepackage[caption=false,font=footnotesize]{subfig}
\usepackage{microtype}
\usepackage{cite}
\usepackage{flushend}

\makeatletter
\let\NAT@parse\undefined
\makeatother

\usepackage{url}

\usepackage[pdfencoding=auto, hidelinks]{hyperref} % Usually, this is the last package to be imported
\usepackage[hang,flushmargin]{footmisc}

\usepackage[capitalize]{cleveref}
\crefname{section}{Sec.}{Secs.}
\Crefname{section}{Section}{Sections}
\Crefname{table}{Table}{Tables}
\crefname{table}{Tab.}{Tabs.}

\newcommand{\greyrule}{\arrayrulecolor{black!30}\midrule\arrayrulecolor{black}}

\definecolor{Gray}{gray}{0.9}

\newcommand{\cmark}{\ding{51}}  % pick from here: https://i.stack.imgur.com/Vjm6r.png

\begin{document}

\title{\LARGE \bf
Few-Shot Panoptic Segmentation With Foundation Models
}

\author{
Markus Käppeler$^{1*}$,
Kürsat Petek$^{1*}$,
Niclas Vödisch$^{1*}$,
Wolfram Burgard$^{2}$,
and Abhinav Valada$^{1}$% <-this % stops a space
\thanks{$^{*}$ Equal contribution.}%
\thanks{$^{1}$ Department of Computer Science, University of Freiburg, Germany.}%
\thanks{$^{2}$ Department of Eng., University of Technology Nuremberg, Germany.}%
\thanks{This work was funded by the German Research Foundation (DFG) Emmy Noether Program grant No 468878300 and the European Union’s Horizon 2020 research and innovation program grant No 871449-OpenDR.}%
\thanks{Accepted for the 2024 IEEE Int. Conf. on Robotics and Automation.}
}

\maketitle

%%%%%%%%%%%%%%%%%%%%%%%%%%%%%%%%%%%%%%%%%%%%%%%%%%%%%%%%%%%%%%%%%%%%%%%%%%%%%%%%

\begin{abstract}
    Current state-of-the-art methods for panoptic segmentation require an immense amount of annotated training data that is both arduous and expensive to obtain posing a significant challenge for their widespread adoption. Concurrently, recent breakthroughs in visual representation learning have sparked a paradigm shift leading to the advent of large foundation models that can be trained with completely unlabeled images.
In this work, we propose to leverage such task-agnostic image features to enable few-shot panoptic segmentation by presenting Segmenting Panoptic Information with Nearly 0 labels (SPINO). In detail, our method combines a DINOv2 backbone with lightweight network heads for semantic segmentation and boundary estimation. We show that our approach, albeit being trained with only ten annotated images, predicts high-quality pseudo-labels that can be used with any existing panoptic segmentation method. Notably, we demonstrate that SPINO achieves competitive results compared to fully supervised baselines while using less than 0.3\% of the ground truth labels, paving the way for learning complex visual recognition tasks leveraging foundation models. To illustrate its general applicability, we further deploy SPINO on real-world robotic vision systems for both outdoor and indoor environments. To foster future research, we make the code and trained models publicly available at \mbox{\url{http://spino.cs.uni-freiburg.de}}.

\end{abstract}

%%%%%%%%%%%%%%%%%%%%%%%%%%%%%%%%%%%%%%%%%%%%%%%%%%%%%%%%%%%%%%%%%%%%%%%%%%%%%%%%

\section{Introduction}

Panoptic segmentation~\cite{kirillov2019panoptic} poses an important contribution to holistic scene understanding by enabling robots to assign semantic meaning to their environment while delineating individual objects. However, most previous methods addressing panoptic segmentation rely on supervised training~\cite{cheng2020panoptic, mohan2022perceiving}, hence requiring a large amount of ground truth labels. This hinders their widespread adoption as generating panoptic annotations is both expensive and time-consuming, e.g., manually labeling a single high-resolution image of urban scenarios takes approximately \SI{1.5}{\hour}~\cite{cordts2016cityscapes}. Therefore, it is paramount to reduce the number of required labels~\cite{chen2020naivestudent}, e.g., by advancing weakly- and unsupervised methods or by leveraging task-agnostic pretraining strategies~\cite{lang2023self}.

Facing similar issues, the domain of natural language processing (NLP) has recently seen a rise of large foundation models~\cite{touvron2023llama}. This paradigm shift in NLP also inspired the vision community to propose similar methods such as CLIP~\cite{radford2021clip} or Segment Anything~\cite{kirillov2023sam}. While both still require some supervision signal, e.g., from image captions or coarse object masks, DINO~\cite{caron2021dino} learns visual representation in a fully unsupervised manner allowing to significantly extend the amount of usable resources. Prior works have shown that one can bootstrap such general representations for several tasks including depth estimation~\cite{oquab2023dinov2}, semantic segmentation~\cite{oquab2023dinov2, hamilton2022stego}, and object detection~\cite{wang2023cut}.
Based on these findings, we argue that it is time for a fundamental paradigm switch for vision tasks that exploit task-agnostic foundation models to enable few-shot training. In contrast to unsupervised techniques~\cite{hamilton2022stego, hyun2021picie}, we show that such an approach can yield results competitive with fully supervised learning methods.\looseness=-1

\begin{figure}[t]
    \centering
    \includegraphics[width=\linewidth]{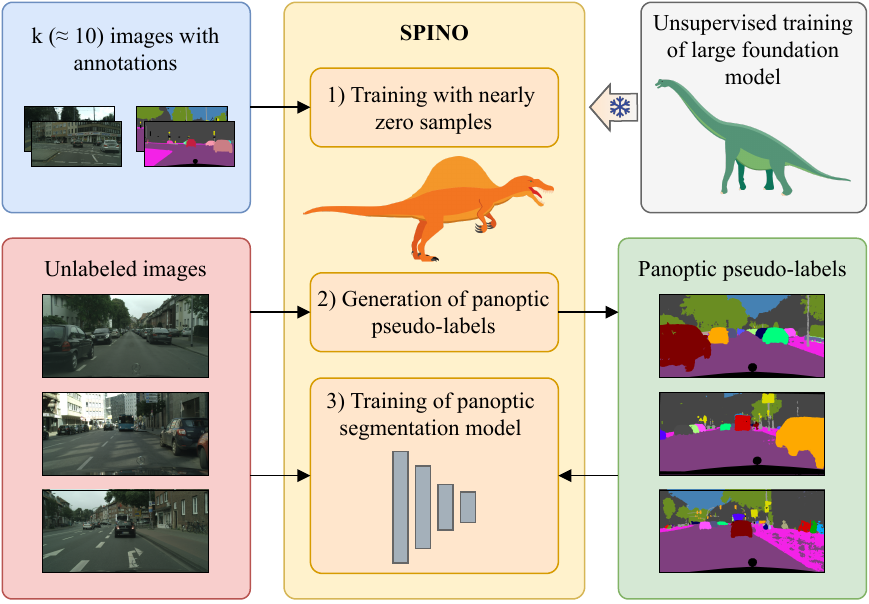}
    \vspace{-.5cm}
    \caption{SPINO enables few-shot panoptic segmentation by exploiting descriptive image features from unsupervised task-agnostic pretraining. We generate panoptic pseudo-labels by learning from only $k \approx 10$ annotated images in an offline manner. We can then leverage these pseudo-labels to train any panoptic segmentation model enabling online deployment.}
    \label{fig:cover}
    \vspace{-.2cm}
\end{figure}

In this work, we present a method for \textit{Segmenting Panoptic Information with Nearly 0 labels} (SPINO), illustrated in \cref{fig:cover}. First, we leverage a frozen DINOv2~\cite{oquab2023dinov2} backbone to extract visual features. Subsequently, we train two task-specific heads for semantic segmentation and boundary estimation with as few as ten annotated images to perform few-shot panoptic segmentation. To enable real-time inference and to further boost the quality of our predictions, we generate panoptic pseudo-labels in an offline manner for a larger bag of raw images that can then be used to train any existing panoptic segmentation model.
We perform extensive evaluations on several public~\cite{cordts2016cityscapes, liao2022kitti360} and in-house datasets that demonstrate that our SPINO approach yields results that are highly competitive with fully supervised learning models. In particular, our extensive evaluations suggest that few-shot panoptic segmentation provides the means to soon become on par with supervised state-of-the-art methods.

To summarize, the main contributions are as follows:
\begin{enumerate}[topsep=0pt]
    \item We propose the first method for few-shot panoptic segmentation based on unsupervised foundation models.
    \item We present a novel pseudo-label generation scheme that can be trained with as few as ten annotated images.
    \item We show that SPINO yields results that are competitive to supervised training with ground truth labels.
    \item In extensive evaluations, we illustrate the effect of various architectural design choices and apply our method to real-world robotic vision platforms.
    \item We make the code and trained models publicly available at \mbox{\url{http://spino.cs.uni-freiburg.de}}.
\end{enumerate}

\section{Related Work}
\label{sec:related-work}

In this section, we present an overview of panoptic segmentation, visual representation learning, and both unsupervised and weakly-supervised image segmentation techniques.\looseness=-1

%%%%%%%%%%%%%%%%%%%%%%%%%%%%%%%%%%%%%%%%%%%%%%%%%%%%%%%%%%%%%%%%%%%%%%%%%%%%%%%%

{\parskip=3pt
\noindent\textit{Panoptic Segmentation:}
Panoptic segmentation~\cite{kirillov2019panoptic} combines semantic and instance segmentation into a single task with two categories of scene elements. The static background comprises the so-called ``stuff'' classes such as \textit{buildings}, whereas dynamic objects such as \textit{cars} belong to the ``thing'' category. While ``stuff'' classes only receive a semantic label, ``thing'' classes are further separated on an instance level.
Since the introduction of this task, several deep learning-based methods~\cite{xiong2019upsnet, porzi2019seamless, mohan2021efficientps, cheng2020panoptic, wang2020axialdeeplab} have been proposed requiring a large amount of data for training.
Recently, the focus has shifted towards more challenging variants, e.g., open-vocabulary methods such as from Ding \textit{et~al.}~\cite{ding2023openvocabulary} leveraging insights from foundation models~\cite{radford2021clip}. Removing the need for labels, CoDEPS~\cite{voedisch23codeps} addresses unsupervised domain adaptation from a source to a previously unseen target domain.
In this work, we propose a method for few-shot panoptic segmentation requiring as few as ten annotated images.
}

%%%%%%%%%%%%%%%%%%%%%%%%%%%%%%%%%%%%%%%%%%%%%%%%%%%%%%%%%%%%%%%%%%%%%%%%%%%%%%%%

{\parskip=3pt
\noindent\textit{Visual Representation Learning:}
Breakthroughs in natural language processing (NLP)~\cite{touvron2023llama} have shown that task-agnostic pretraining can yield feature representations that, fine-tuned to specific applications, become competitive with prior state-of-the-art methods~\cite{brown2020language}.
A common approach to obtaining similar representations in the visual domain is contrastive learning~\cite{he2020moco}. However, although not using human annotations, the choice of the dataset still introduces a significant bias on the learned representation that can be mitigated by extensive data augmentation~\cite{gansbeke2021revisiting}.
Masked autoencoders (MAE)~\cite{he2022masked} represent another type of self-supervised learners that learn to reconstruct areas in an image that have been masked. After pretraining, MAEs can be fine-tuned for various downstream tasks.
More recently, the usage of foundation models in NLP has also started to influence computer vision.
For instance, CLIP~\cite{radford2021clip} leverages insights from constrastive learning by exploiting textual supervision to guide the learning of visual features. However, this text-guided supervision strategy limits the choice of training data.
SAM~\cite{kirillov2023sam} removes the need for captions and relies on a self-iterative training scheme starting from coarse object masks. While showing impressive zero-shot performance for semantic segmentation on unseen domains, it lacks the ability to assign class labels to the segments.
Finally, DINO~\cite{caron2021dino} represents a new family of foundation models that can be trained only from raw images. In particular, DINO demonstrates that such unsupervised pretraining can achieve even more explicit features for semantic segmentation than their supervised counterparts. Further advances have been shown by DINOv2~\cite{oquab2023dinov2} that combines several prior insights with training on a curated dataset.
In this work, we exploit descriptive image features from a DINOv2 backbone to generate panoptic pseudo-labels.
}

%%%%%%%%%%%%%%%%%%%%%%%%%%%%%%%%%%%%%%%%%%%%%%%%%%%%%%%%%%%%%%%%%%%%%%%%%%%%%%%%

{\parskip=3pt
\noindent\textit{Unsupervised and Weakly-Supervised Segmentation:}
Since obtaining pixel-wise annotations for supervised training of image segmentation tasks is expensive, in the last few years research has shifted towards reducing the number of human annotations. Recent methods build on the observation that features from unsupervised pretraining are semantically consistent across images from differing domains~\cite{hamilton2022stego}.
For instance, LOST~\cite{simeoni2021lost} uses DINO~\cite{caron2021dino} features for bounding box extraction to bootstrap supervised training of an object detector. Objects can be assigned to the same class via $k$-means clustering in the feature space.
Similarly, TokenCut~\cite{wang2023tokencut} relies on Normalized Cut (NCut)~\cite{shi2000normalized} to group self-similar image regions based on DINO features.
While these previous methods work well for foreground/background segmentation, FreeSOLO~\cite{wang2022freesolo} addresses multi-object detection by enhancing coarse masks via one-stage self-training in a weakly supervised manner. However, requiring in-domain data results in a lack of generalization.
In contrast, CutLER~\cite{wang2023cut} achieves impressive zero-shot performance leveraging DINO features to generate coarse masks followed by weakly supervised training of a separate instance segmentation network. Although applicable to multi-object scenarios, relying on iterative NCut requires specifying the number of expected objects.\looseness=-1

With respect to semantic segmentation, MaskContrast~\cite{gansbeke2021unsupervised} and PiCIE~\cite{hyun2021picie} are notable methods from before the advent of large pretraining models. While MaskContrast contrasts learned features within and across saliency masks, PiCIE searches for descriptive image features guided by photometric invariance and geometric equivariance.
Recently, both MaskDistill~\cite{gansbeke2022discovering} and STEGO~\cite{hamilton2022stego} leverage features from a frozen DINO~\cite{caron2021dino} backbone. To further refine the pretrained features, STEGO adds a task-specific segmentation head followed by clustering.
Other examples of exploiting foundation models include CLIP-ES~\cite{lin2023clipes}, which relies on contrastive language-image pretraining~\cite{radford2021clip}, and SEPL~\cite{chen2023segment} that combines the class-agnostic masks from SAM~\cite{kirillov2023sam} with class activation maps for class assignment.
To the best of our knowledge, our proposed SPINO constitutes the first attempt to directly exploit fully unsupervised representation pretraining for panoptic segmentation.
}

\begin{figure*}[t]
    \centering
    \includegraphics[width=\linewidth]{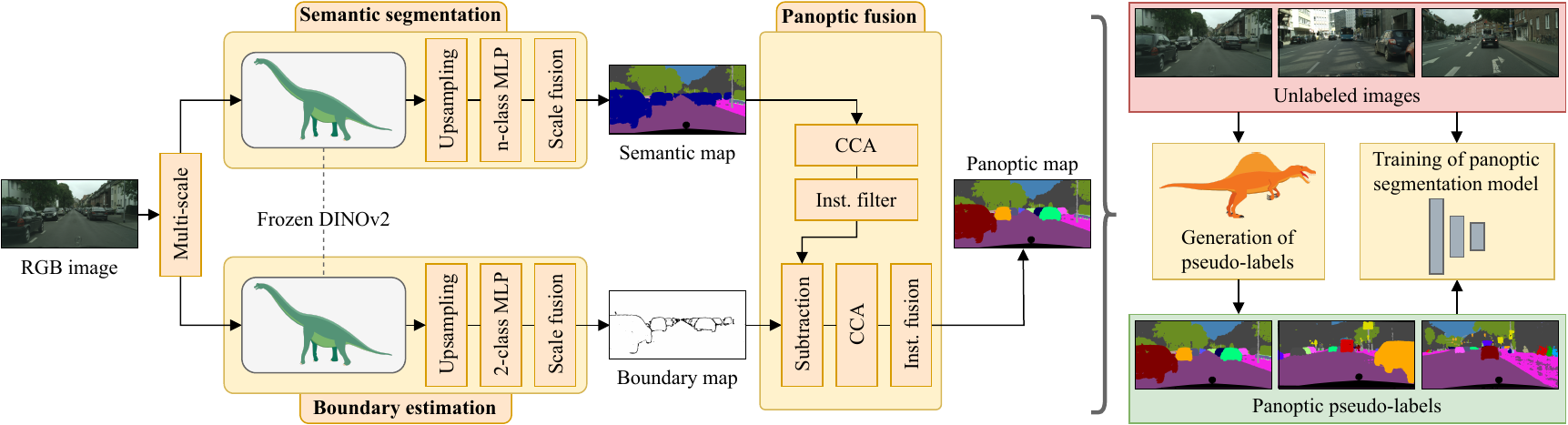}
    \vspace*{-.5cm}
    \caption{Overview of our proposed SPINO approach for few-shot panoptic segmentation. SPINO consists of two learning-based modules for semantic segmentation and boundary estimation that leverage features from the recent foundation model DINOv2~\cite{oquab2023dinov2}. A panoptic fusion scheme combines their outputs using connected component analysis (CCA) and multiple small instance filtering steps. SPINO creates pseudo-labels for a large number of unlabeled images using only $k \approx 10$ images with ground truth annotations. These pseudo-labels can then be utilized to train any panoptic segmentation model.}
    \label{fig:overview}
\end{figure*}

\section{Technical Approach}

In this section, we present our proposed approach SPINO for few-shot panoptic segmentation. As illustrated in \cref{fig:overview}, we leverage the recent foundation model DINOv2~\cite{oquab2023dinov2} to extract descriptive image features for both semantic segmentation and boundary estimation. In particular, we propose a novel pseudo-label generation scheme that separates semantic regions of ``thing'' classes into individual instances by predicting object boundaries. With this approach, SPINO can bootstrap very few ground truth annotations for generating high-quality panoptic pseudo-labels. To enable real-time inference and to further boost the quality of our panoptic predictions, we train a panoptic segmentation model using the generated pseudo-labels.

%%%%%%%%%%%%%%%%%%%%%%%%%%%%%%%%%%%%%%%%%%%%%%%%%%%%%%%%%%%%%%%%%%%%%%%%%%%%%%%%

\subsection{Few-Shot Pseudo-Label Generation}
\label{ssec:ta-label-generator}

We propose a novel panoptic segmentation scheme to generate panoptic pseudo-labels in an offline manner while requiring very few ground truth annotations for training. Our label generator consists of three main building blocks shown in \cref{fig:overview}, namely learnable modules for semantic segmentation and boundary estimation as well as a static component to fuse their predictions.
The semantic segmentation module is comprised of a frozen DINOv2~\cite{oquab2023dinov2} backend, a bilinear \mbox{14x-upsampling} layer, and a final \mbox{$n$-class} MLP with 4 layers. Here, $n$ denotes the number of semantic classes as specified in \cref{ssec:datasets}. In detail, we use the DINOv2 weights of the ViT-B/14 variant provided by the authors. For the boundary estimation module, we employ a similar design but use \mbox{4x-upsampling} and set $n=2$ for binary classification. 

%%%%%%%%%%%%%%%%%%%%%%%%%%%%%%%%%%%%%%%%%%

\begin{figure}
    \centering
    \includegraphics[width=\linewidth]{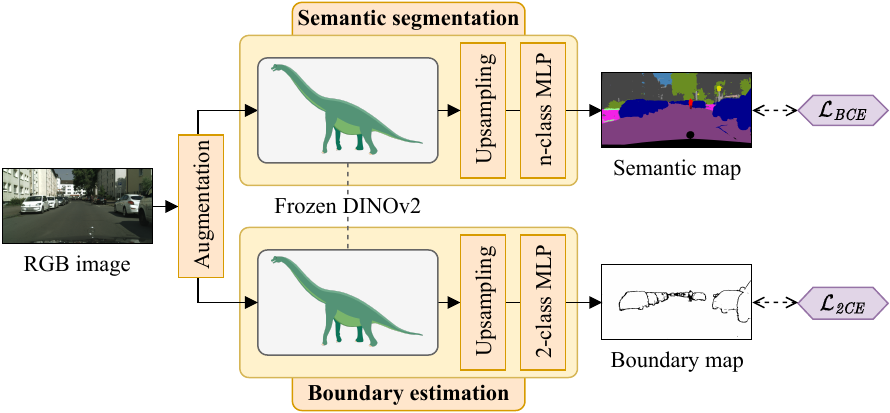}
    \vspace{-.5cm}
    \caption{Our proposed pseudo-label generator comprises two learnable modules for semantic segmentation and boundary estimation that exploit descriptive image features from the recent DINOv2~\cite{oquab2023dinov2} foundation model, enabling training with only $k \approx 10$ ground truth panoptic annotations.}
    \label{fig:panoptic-training}
\end{figure}

{\parskip=3pt
\noindent\textit{Training the Label Generator:}
A key idea of SPINO is to train our proposed pseudo-label generator with only $k$ ground truth annotations, where $k$ denotes numbers as small as 10. Notably, the unsupervised training procedure of DINOv2 does not further increase this number even when considering the pretraining.
We illustrate the training of our pseudo-label generator in \cref{fig:panoptic-training}. First, to stabilize the training with such few samples, we employ various data augmentation techniques on the input RGB image including random cropping, horizontal flipping, and color jitter. Subsequently, we feed the augmented image to the two task-specific heads and compute the respective loss functions.

We supervise the semantic segmentation head with the bootstrapped cross-entropy loss function $\mathcal{L}_\mathit{BCE}$~\cite{pohlen2017full} to account for rare classes.
\begin{equation}
    \mathcal{L}_\mathit{BCE} = - \frac{1}{K} \sum_{i=1}^N \mathds{1} \left[ p_{i, y_i} < t_K \right] \cdot \log (p_{i, y_i}) \, ,
    \label{eqn:bce-loss}
\end{equation}
where $p_{i, y_{i}}$ denotes the posterior probability of pixel~$i \in [1, N]$ for its ground truth class~$y_i \in \{1, ..., c\}$ with $N$ and $c$ being the number of pixels and classes, respectively. The indicator function~$\mathds{1}(\cdot)$ is 1 if $p_{i, y_i}$ is below a threshold $t_K$ and 0 otherwise. Following previous works~\cite{cheng2020panoptic, voedisch23codeps}, we set $t_K = 0.2$ such that only those pixels with top-K highest losses contribute to $\mathcal{L}_\mathit{BCE}$.
In order to train the boundary estimation module, we generate ground truth boundary maps as follows: If the instance ID of a pixel is different from any of its eight neighbors, we assign 1 to this pixel. Otherwise, we set the value of the center pixel to 0. During training, we compute the binary cross-entropy loss $\mathcal{L}_\mathit{2CE}$ as the supervision signal.
\begin{equation}
    \mathcal{L}_\mathit{2CE} = - \frac{1}{N} \sum_{i=1}^N y_i \cdot \log (p_{i}) + (1 - y_i) \cdot \log (1 - p_{i}) \, ,
\end{equation}
where $y_i \in \{0, 1\}$ is the binary boundary label of pixel~$i$ and $p_{i}$ denotes the probability of the pixel $i$ being a boundary.
}

%%%%%%%%%%%%%%%%%%%%%%%%%%%%%%%%%%%%%%%%%%

{\parskip=3pt
\noindent\textit{Employing the Label Generator:}
In the next step, we leverage the aforementioned trained modules for semantic segmentation and boundary estimation to generate panoptic pseudo-labels for a large number of unlabeled images. In the following, we describe the procedure as depicted in \cref{fig:overview}. Inspired by ensemble learning, we use multi-scale test-time augmentation for both semantic segmentation and boundary estimation. For instance, for scale $s=2$, we divide the image into four equally sized regions, upsample each region to the size of the original input image ($s=1$), and obtain their softmax features. In the scale fusion block, we downsample these feature maps to the original size of the region, join the features of all regions in a single $s=1$ map, and compute the mean across the considered scales. In detail, we use scales $\{1, 2, 3\}$ for the semantic head and scales $\{3, 4, 5\}$ for the boundary estimation head.
Next, we feed the predicted semantic map and the estimated object boundary map to our panoptic fusion module. First, for each ``thing'' class, we perform connected component analysis~(CCA) yielding disconnected blobs. If a blob consists of fewer pixels than a threshold, we assign the semantic \textit{void} class to its pixels. Otherwise, we subtract the predicted border for this blob from the semantic map followed by CCA to detect separate instances within a blob. If the number of pixels of an instance is below another threshold, we add it to its nearest neighbor which fulfills the minimum size requirement. If all instances of a blob are below this threshold, we combine them into a single instance.
Finally, due to the top-down approach, the inferred instance maps already contain semantic information leading to the desired pseudo-labels for panoptic segmentation.
}

%%%%%%%%%%%%%%%%%%%%%%%%%%%%%%%%%%%%%%%%%%%%%%%%%%%%%%%%%%%%%%%%%%%%%%%%%%%%%%%%

\subsection{Training a Panoptic Segmentation Model}
\label{ssec:ta-ps-model}

\begin{figure}[t]
    \centering
    \includegraphics[width=\linewidth]{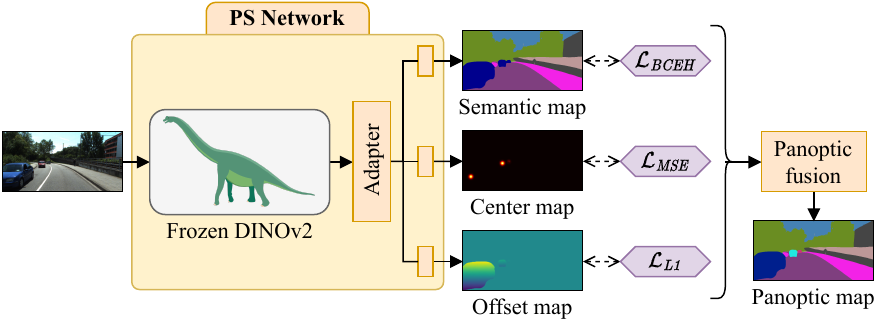}
    \vspace{-.5cm}
    \caption{To enable online predictions and to further boost the performance compared to the pseudo-label generator, we train a bottom-up panoptic segmentation model using our generated pseudo-labels. The network consists of a frozen DINOv2~\cite{oquab2023dinov2} backbone with an adapter~\cite{chen2023vision} and three task-specific heads, whose output is merged by a panoptic fusion module~\cite{cheng2020panoptic}.}
    \label{fig:panoptic-segmentation-model}
\end{figure}

After creating pseudo-labels for a large set of unlabeled images, we train a panoptic segmentation model as illustrated in \cref{fig:overview}. In contrast to the offline label generator, such a model allows for online panoptic segmentation while further enhancing the overall performance. Although this approach is generally applicable to any panoptic segmentation model, in this work, we follow the spirit of our pseudo-label generator. In detail, our bottom-up panoptic segmentation network consists of a frozen DINOv2~\cite{oquab2023dinov2} backbone with an adapter module~\cite{chen2023vision} and three task-specific heads~\cite{cheng2020panoptic} for semantic segmentation, instance center prediction, and pixel offset regression, respectively. In \cref{fig:panoptic-segmentation-model}, we visualize this architecture. The semantic head predicts a semantic class for each pixel and is trained with the bootstrapped cross-entropy loss with hard pixel mining~\cite{cheng2020panoptic}.
\begin{equation}
    \mathcal{L}_\mathit{BCEH} = - \frac{1}{K} \sum_{i=1}^N w_i \cdot \mathds{1} \left[ p_{i, y_i} < t_K \right] \cdot \log (p_{i, y_i}) \, ,
\end{equation}
which builds upon \cref{eqn:bce-loss} but adds weights $w_i > 1$ for pixels that belong to small instances. For other instances and ``stuff'' classes, the pixel weight remains at $w_i = 1$. Similar to \cref{eqn:bce-loss}, we set $t_K = 0.2$.
Addressing instance segmentation, the center head generates a probability map with high values for instance centers and the offset head estimates the 2D offset of a pixel to the nearest instance center. To train these heads, we utilize the MSE loss~$\mathcal{L}_\mathit{MSE}$ for the center head and the L1 loss~$\mathcal{L}_\mathit{L1}$ for the offset head.
Consequently, we compute the total loss as a weighted sum:
\begin{equation}
    \mathcal{L}_\mathit{PAN} = \lambda_\mathit{sem} \mathcal{L}_\mathit{BCEH} + \lambda_\mathit{cen} \mathcal{L}_\mathit{MSE} + \lambda_\mathit{off} \mathcal{L}_\mathit{L1}
\end{equation}

To increase the learning speed, we propose to further exploit the $k$ annotated images, which were used to train the pseudo-label generator, also when training the panoptic segmentation model. In particular, we construct batches that contain both pseudo-labels and one ground truth sample. Formally, a batch $\mathbf{b}$ of size $n$ is given by
\begin{equation}
    \mathbf{b} = \{ \mathbf{\hat{I}_1}, \dots, \mathbf{\hat{I}_{n-1}}, \mathbf{I_\mathit{GT}} \} \, ,
\end{equation}
where $\mathbf{\hat{I}_i}$ denote pseudo-labeled images and $\mathbf{I_\mathit{GT}}$ is from the set of $k$ images with ground truth labels. We further apply data augmentation via color jitter and horizontal flipping.

During test-time, a panoptic fusion module~\cite{cheng2020panoptic} predicts the final panoptic segmentation map from the output of the individual heads, shown in \cref{fig:panoptic-segmentation-model}. In detail, it assigns a semantic label to the class-agnostic instance predictions using majority voting over the semantic predictions of all pixels within an instance.

\section{Experimental Evaluation}

In this section, we demonstrate that our proposed SPINO outperforms unsupervised methods for semantic segmentation and yields competitive results compared to fully supervised setups for panoptic segmentation that require a huge number of ground truth annotations. We provide both quantitative and qualitative results on multiple public and in-house datasets. Finally, we extensively evaluate several design choices for our pseudo-label generator.

\begin{figure*}[t]
    \centering
    \includegraphics[width=\linewidth]{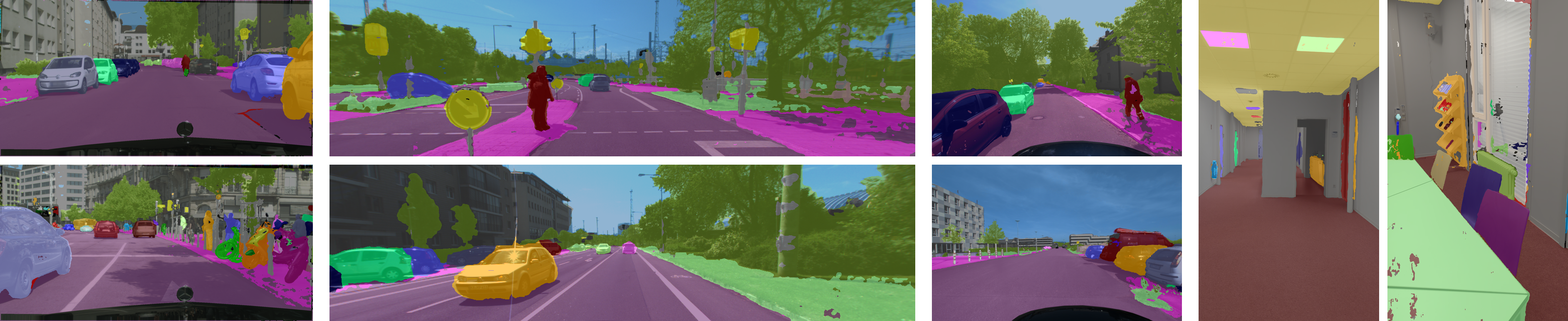}
    \includegraphics[width=\linewidth]{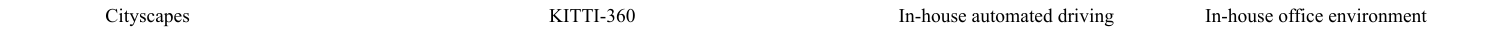}
    \vspace*{-0.7cm}
    \caption{Qualitative performance of our pseudo-label generator in four diverse domains from both public and in-house data sources. From left to right, we show two samples each for Cityscapes~\cite{cordts2016cityscapes}, KITTI-360~\cite{liao2022kitti360}, in-house automated driving, and an in-house office environment.}
    \label{fig:qualitative-results}
    % \vspace{-.3cm}
\end{figure*}

%%%%%%%%%%%%%%%%%%%%%%%%%%%%%%%%%%%%%%%%%%%%%%%%%%%%%%%%%%%%%%%%%%%%%%%%%%%%%%%%

\subsection{Datasets}
\label{ssec:datasets}

We present results on various datasets including the public Cityscapes~\cite{cordts2016cityscapes} and KITTI-360~\cite{liao2022kitti360} as well as our in-house data for automated driving and from an indoor office environment.

%%%%%%%%%%%%%%%%%%%%%%%%%%%%%%%%%%%%%%%%%%

{\parskip=3pt
\noindent\textit{Cityscapes:}
The Cityscapes dataset~\cite{cordts2016cityscapes} contains RGB images and fine panoptic annotations for automated driving in 50 cities across Germany and bordering regions. We select $k$ images from the \textit{train} split to train our label generator and generate pseudo-labels for the remaining images. In a separate experiment, we also generate pseudo-labels on the entire \textit{train\_extra} split. To evaluate the performance, we report metrics on the \textit{val} split. When creating the pseudo-labels, we mask out the hood of the ego car as it remains static and hence can be inferred from the $k$ annotated images~\cite{chen2020naivestudent}. We report metrics using 19 classes as per the official Cityscapes evaluation protocol.
}

%%%%%%%%%%%%%%%%%%%%%%%%%%%%%%%%%%%%%%%%%%

{\parskip=3pt
\noindent\textit{KITTI-360:}
The KITTI-360 dataset~\cite{liao2022kitti360} was recorded in Karlsruhe, Germany, and provides RGB images and panoptic annotations for sequential data. Following prior works~\cite{mohan2022amodal, gosala2023skyeye}, we use sequence 10 for evaluation and the remaining sequences for the pseudo-label generation. We report results using 14 classes as detailed by Vödisch \textit{et~al.}~\cite{voedisch23codeps}.
}

%%%%%%%%%%%%%%%%%%%%%%%%%%%%%%%%%%%%%%%%%%

{\parskip=3pt
\noindent\textit{In-House:}
To illustrate the main benefit of SPINO, i.e., enabling panoptic segmentation on different vision systems with very few reference annotations, we employ our method on two in-house data sources. First, following the spirit of the public datasets, we use an automated driving perception car navigating in Freiburg, Germany. Second, to demonstrate general applicability, we record indoor data in our office environment. For both domains, we prepare annotations for ten images to train the pseudo-label generator.
}

%%%%%%%%%%%%%%%%%%%%%%%%%%%%%%%%%%%%%%%%%%%%%%%%%%%%%%%%%%%%%%%%%%%%%%%%%%%%%%%%

\begin{table}[t]
\scriptsize
\centering
\caption{Panoptic/semantic segmentation on Cityscapes}
\vspace{-0.2cm}
\label{tab:baselines}
\setlength\tabcolsep{3.7pt}
\begin{threeparttable}
    \begin{tabular}{ l | c | ccccc }
        \toprule
        \textbf{Method} & \textbf{Train. data} & Acc & mIoU & PQ & SQ & RQ  \\
        \midrule
        \multicolumn{2}{l}{\textit{Fully supervised}} \\
        [.25ex]
        DINOv2 + Adapt. + PH & GT & 91.9 & 77.0 & 51.4 & 78.9 & 63.1 \\
        \midrule
        \multicolumn{2}{l}{\textit{Unsupervised}} \\
        [.25ex]
        Modified DC~\cite{caron2018mdc} & n/a & 35.3 & 6.8 & -- & -- & -- \\
        PiCIE~\cite{hyun2021picie} & n/a & 72.7 & 13.8 & -- & -- & -- \\
        STEGO~\cite{hamilton2022stego} & n/a & 89.1 & 38.0 & -- & -- & -- \\
        \midrule
        \multicolumn{2}{l}{\textit{Few-shot supervision}} \\
        [.25ex]
        ResNet-50 + PH & 10 GT & 74.9 & 32.1 & 16.8 & 45.6 & 20.8 \\
        DINOv2 + PH & 10 GT & 81.6 & 49.4 & 20.6 & 49.9 & 25.8 \\
        DINOv2 + Adapt. + PH & 10 GT & 82.8 & 52.5 & 22.0 & 60.9 & 27.0 \\
        \greyrule
        Pseudo-labels (\textit{ours}) & 10 GT & 86.0 & 61.5 & 35.9 & 73.7 & 45.9 \\
        SPINO (\textit{ours}) & PL & 86.3 & 60.6 & 36.4 & 73.5 & 46.7 \\
        \hspace{.5pt} + Mixed-batch & PL & 86.6 & 61.2 & 36.5 & 74.8 & 46.3 \\
        SPINO (\textit{ours}) & PL++ & 86.6 & 61.8 & 37.2 & 74.5 & 47.5 \\
        \bottomrule
    \end{tabular}
    \footnotesize
    PH refers to the panoptic heads as shown in \cref{fig:panoptic-segmentation-model}. GT and PL indicate training with ground truth annotations and pseudo-labels, where the ``PL++'' marks pseudo-labels on the \textit{train\_extra} split. The architecture of SPINO corresponds to ``DINOv2 + Adapt. + PH''.
    \vspace{-.2cm}
\end{threeparttable}
\end{table}

\subsection{Panoptic Segmentation}
\label{ssec:exp-panoptic-segmentation}

To evaluate the performance of SPINO, we measure the pixel accuracy (Acc) and the mean IoU (mIoU) for semantic segmentation as well as the panoptic quality (PQ), the segmentation quality (SQ), and the recognition quality (RQ) for panoptic segmentation. Based on the ablation studies in \cref{ssec:ablation-studies}, we train our pseudo-label generator on $k=10$ human-selected, labeled images with a batch size $b=1$ and a learning rate $\mathit{lr}=0.001$.

%%%%%%%%%%%%%%%%%%%%%%%%%%%%%%%%%%%%%%%%%%

{\parskip=3pt
\noindent\textit{Few-Shot Training:}
First, we illustrate the efficacy of our pseudo-label generation scheme. As shown by the metrics in \cref{tab:baselines}, training Panoptic-DeepLab~\cite{cheng2020panoptic} (with a \mbox{ResNet-50} backbone) on only ten images yields poor results that can be improved by replacing the backbone with a frozen DINOv2~\cite{oquab2023dinov2}. Following the common methodology for dense prediction tasks, we also add an adapter module~\cite{chen2023vision} to further increase the performance. However, the results remain significantly inferior to the quality of our pseudo-labels with respect to both semantic and panoptic segmentation. Notably, our pseudo-label generator comprises a much simpler design, e.g., estimating object boundaries instead of predicting instance centers and pixel offsets.
For the overall SPINO approach, we adopt the network design of DINOv2 plus an adapter module. Naive training on the generated pseudo-labels already yields highly competitive results compared to training with ground truth labels considering that we use less than \SI{0.29}{\percent} of the labels. We further show how the proposed mixed-batch strategy that closely incorporates the ten ground truth labels increases all three semantic metrics.
\looseness=-1

Next, we also generate pseudo-labels for the unlabeled \textit{train\_extra} split of Cityscapes, increasing the amount of training data for the panoptic segmentation model. The results in \cref{tab:baselines} indicate that our approach opens up an avenue for exploiting unlabeled large-scale data recordings for the training of existing panoptic segmentation methods.
}

%%%%%%%%%%%%%%%%%%%%%%%%%%%%%%%%%%%%%%%%%%

{\parskip=3pt
\noindent\textit{Comparison with Unsupervised Segmentation:}
Second, we compare SPINO to the state-of-the-art for unsupervised semantic segmentation. As we follow the official Cityscapes evaluation protocol, we retrain PiCIE~\cite{hyun2021picie} and their modified DeepCluster~\cite{caron2018mdc, hyun2021picie} using the released code on 19 classes. For STEGO~\cite{hamilton2022stego}, we use the provided network weights but reevaluate on 19 classes. Note that, for both PiCIE and STEGO, reducing the number of classes leads to higher metrics than reported by the authors. As SPINO significantly outperforms these baselines, we argue that requiring ten instead of zero annotated images is well justified.
}

%%%%%%%%%%%%%%%%%%%%%%%%%%%%%%%%%%%%%%%%%%

\begin{table*}[t]
\scriptsize
\begin{minipage}{.62\linewidth}
    \centering
    \caption{Panoptic segmentation on Cityscapes and KITTI-360}
    \vspace{-0.2cm}
    \label{tab:quantitative-results}
    \setlength\tabcolsep{3.8pt}
    \begin{threeparttable}
        \begin{tabular}{ l | c | ccccc | ccccc }
            \toprule
             & \textbf{Train.} & \multicolumn{5}{c|}{Cityscapes} & \multicolumn{5}{c}{KITTI-360} \\
            \textbf{Method} & \textbf{data} & Acc & mIoU & PQ & SQ & RQ & Acc & mIoU & PQ & SQ & RQ  \\
            \midrule
            Pseudo-labels & 10 GT & 86.0 & 61.5 & 35.9 & 73.7 & 45.9 & 75.8 & 54.7 & 32.5 & 70.7 & 42.1 \\
            \midrule
            ResNet-50 + PH & GT & 89.4 & 64.9 & 44.2 & 75.3 & 56.1 & 83.0 & 64.1 & 41.0 & 76.5 & 50.5 \\
            DINOv2 + PH & GT & 89.4 & 71.4 & 41.0 & 74.4 & 51.7 & 83.5 & 62.8 & 39.3 & 70.5 & 48.7 \\
            DINOv2 + Adapt. + PH & GT & 91.9 & 77.0 & 51.4 & 78.9 & 63.1 & 86.0 & 65.6 & 42.5 & 72.9 & 51.2 \\
            \midrule
            ResNet-50 + PH & PL & 85.4 & 57.3 & 33.0 & 67.8 & 42.3 & 76.2 & 52.1 & 32.2 & 67.6 & 41.0\\ 
            DINOv2 + PH & PL & 84.5 & 57.1 & 31.4 & 70.9 & 40.3 & 76.4 & 54.6 & 32.7 & 71.7 & 42.0  \\
            \rowcolor{Gray}
            DINOv2 + Adapt. + PH & PL & 86.3 & 60.6 & 36.4 & 73.5 & 46.7 & 76.6 & 55.5 & 33.3 & 71.9 & 42.8   \\
            \bottomrule
        \end{tabular}
        \footnotesize
        PH refers to the panoptic heads shown in \cref{fig:panoptic-segmentation-model}. GT and PL indicate ground truth annotations and pseudo-labels. The gray row corresponds to SPINO without mixed-batch training.
    \end{threeparttable}
\end{minipage}
\hfill
\begin{minipage}{.36\linewidth}
    \centering
    \caption{Ablation study: network architecture}
    \vspace{-0.2cm}
    \label{tab:ablation-architecture}
    \setlength\tabcolsep{2.8pt}
    \begin{threeparttable}
        \begin{tabular}{l | ccccc | ccccc}
            \toprule
            \textbf{Method} & \rotatebox{90}{A: k-NN} & \rotatebox{90}{B: Lin. Layer} & \rotatebox{90}{C: CNN} & \rotatebox{90}{D: MLP} & \rotatebox{90}{E: Upsamling} & Acc & mIoU & PQ & SQ & RQ \\
            \midrule
            A & \cmark &  & & &  & 78.7 & 51.7 & 26.1 & 68.5 & 35.0 \\
            B & & \cmark & & &  & 84.3 & 60.0 & 32.6 & 71.3 & 42.5 \\
            B + E & & \cmark & & & \cmark & 84.3 & 60.0 & 33.6 & 71.7 & 43.8 \\
            C + E & & & \cmark & & \cmark & 82.9 & 55.1 & 29.7 & 70.9 & 38.4 \\
            \rowcolor{Gray}
            D + E & & & & \cmark & \cmark & \textbf{86.0} & \textbf{61.5} & \textbf{35.9} & \textbf{73.7} & \textbf{45.9} \\
            \bottomrule
        \end{tabular}
        \footnotesize
        Due to the high computational complexity, the k-NN is evaluated without training data augmentation.
    \end{threeparttable}
\end{minipage}
% \vspace*{-.3cm}
\end{table*}

\begin{table*}
\scriptsize
\begin{minipage}{.4\linewidth}
    \centering
    \caption{Ablation study: data augmentation}
    \vspace{-0.2cm}
    \label{tab:ablation-data-augmentations}
    \setlength\tabcolsep{3.5pt}
    \begin{threeparttable}
        \begin{tabular}{l | ccccc}
            \toprule
            \textbf{Method} & Acc & mIoU & PQ & SQ & RQ \\
            \midrule
            Base & 83.2 & 55.8 & 29.5 & 70.8 & 38.0 \\
            \greyrule
            \multicolumn{2}{l}{\textit{Training time}} \\
            [.25ex]
            \hspace{.5pt} + Random flip & 83.3 & 56.1 & 29.5 & 70.8 & 38.0 \\
            \hspace{.5pt} + Random crop & 83.0 & 57.2 & 30.0 & 70.7 & 39.1 \\
            \hspace{.5pt} + Color jitter & 83.1 & 57.3 & 30.1 & 70.9 & 39.1 \\    
            \greyrule
            \multicolumn{2}{l}{\textit{Test time}} \\
            [.25ex]
            \rowcolor{Gray}
            \hspace{.5pt} + Multi-scale ensemble & \textbf{86.0} & \textbf{61.5} & \textbf{35.9} & \textbf{73.7} & \textbf{45.9} \\
            \bottomrule
        \end{tabular}
        \footnotesize
    \end{threeparttable}
\end{minipage}
\hfill
\begin{minipage}{.28\linewidth}
    \centering
    \caption{Ablation study: batch size}
    \vspace{-0.2cm}
    \label{tab:ablation-batch-size}
    \setlength\tabcolsep{3.5pt}
    \begin{threeparttable}
        \begin{tabular}{c | ccccc}
            \toprule
            \textbf{\makecell{Batch \\ size}} & Acc & mIoU & PQ & SQ & RQ \\
            \midrule
            \rowcolor{Gray}
            1 & \textbf{86.0} & \textbf{61.5} & \textbf{35.9} & \textbf{73.7} & \textbf{45.9} \\
            2 & 84.9 & 59.8 & 34.0 & 72.6 & 43.7 \\    
            4 & 85.3 & 59.4 & 33.6 & 72.3 & 43.3 \\
            8 & 84.5 & 56.8 & 31.3 & 71.4 & 39.9 \\
            \bottomrule
        \end{tabular}
        \footnotesize
    \end{threeparttable}
\end{minipage}
\hfill
\begin{minipage}{.3\linewidth}
    \centering
    \caption{Ablation study: number of labels}
    \vspace{-0.2cm}
    \label{tab:ablation-label-count}
    \setlength\tabcolsep{3.5pt}
    \begin{threeparttable}
        \begin{tabular}{c | ccccc}
            \toprule
            \textbf{\makecell{Label \\ count}} & Acc & mIoU & PQ & SQ & RQ \\
            \midrule
            1 & 69.8 & 37.1 & 19.8 & 55.4 & 27.2 \\
            3 & 81.8 & 49.3 & 30.3 & 64.3 & 38.8 \\
            5 & 82.8 & 55.0 & 32.1 & 65.5 & 41.3 \\
            \rowcolor{Gray}
            10 & 86.0 & 61.5 & 35.9 & 73.7 & 45.9 \\
            25 & 88.5 & 66.9 & 39.6 & 74.9 & 50.1 \\    
            50 & 89.4 & 69.1 & 40.9 & 74.8 & 51.6 \\
            100 & 90.3 & 71.3 & 42.9 & 76.3 & 53.8 \\
            \bottomrule
        \end{tabular}
        \footnotesize
    \end{threeparttable}
\end{minipage}
\vspace*{-.3cm}
\end{table*}

%%%%%%%%%%%%%%%%%%%%%%%%%%%%%%%%%%%%%%%%%%

{\parskip=3pt
\noindent\textit{Generalizability:}
Finally, we extend the evaluation to multiple datasets. In \cref{tab:quantitative-results}, we report quantitative results on both Cityscapes~\cite{cordts2016cityscapes} and KITTI-360~\cite{liao2022kitti360}. In detail, we compare supervised training with ground truth annotations to our few-shot approach. Similar to \cref{tab:baselines}, we report results for three backbones, namely ResNet-50~\cite{he2016resnet}, DINOv2~\cite{oquab2023dinov2}, and DINOv2 with an adapter~\cite{chen2023vision}. Considering that our pseudo-labels are generated based on only ten images, the few-shot methods yield impressive results across the board. Note that ten images correspond to \SI{0.29}{\percent} and \SI{0.02}{\percent} of the utilized ground truth labels for Cityscapes and KITTI-360, respectively. Finally, we provide qualitative visualizations of our pseudo-labels in \cref{fig:qualitative-results} for both public datasets as well as our in-house data including outdoor urban and indoor office environments. Further examples are shown in the supplementary video on the project website.
}

%%%%%%%%%%%%%%%%%%%%%%%%%%%%%%%%%%%%%%%%%%%%%%%%%%%%%%%%%%%%%%%%%%%%%%%%%%%%%%%%

\subsection{Ablation Studies of Pseudo-Label Generation}
\label{ssec:ablation-studies}

We extensively evaluate the architectural design of our pseudo-label generator and demonstrate its efficacy in contrast to several alternatives. In \cref{tab:ablation-architecture,,tab:ablation-data-augmentations,,tab:ablation-batch-size,,tab:ablation-label-count}, we highlight the utilized variant in gray.

%%%%%%%%%%%%%%%%%%%%%%%%%%%%%%%%%%%%%%%%%%

{\parskip=3pt
\noindent\textit{Network Architecture:}
In \cref{tab:ablation-architecture}, we compare the architectural design of our pseudo-label generator using MLPs to other network architectures. Similar to other methods~\cite{caron2021dino, simeoni2021lost}, we use a $k$-NN classifier on the DINOv2 feature patches with $k=5$. Due to the high computational complexity of this approach, we omit training data augmentation for the $k$-NN. Next, we utilize a linear layer with and without prior upsampling. Compared to the $k$-NN, these learnable methods yield a significant improvement but remain inferior to the MLPs. Finally, we demonstrate that our design also outperforms a 4-layer CNN with $3\times3$ convolutions.
}

%%%%%%%%%%%%%%%%%%%%%%%%%%%%%%%%%%%%%%%%%%

{\parskip=3pt
\noindent\textit{Data Augmentation:}
Next, we gradually activate the data augmentation techniques and list the results in \cref{tab:ablation-data-augmentations}. Utilizing data augmentation during the training enhances mIoU, PQ, and RQ, whereas the accuracy and SQ remain stable. Additionally, our employed test-time augmentation based on multi-scale ensemble prediction vastly improves the metrics across the board.
}

%%%%%%%%%%%%%%%%%%%%%%%%%%%%%%%%%%%%%%%%%%

{\parskip=3pt
\noindent\textit{Batch Size:}
In \cref{tab:ablation-batch-size}, provide results for various batch sizes. Note that we scale the learning rate proportionally to the batch size and keep the number of epochs constant. Due to leading to the highest quality of the pseudo-labels, we select a batch size $b = 1$.
}

%%%%%%%%%%%%%%%%%%%%%%%%%%%%%%%%%%%%%%%%%%

{\parskip=3pt
\noindent\textit{Number of Ground Truth Labels:}
Finally, we investigate the effect of the label count on the quality of the pseudo-labels. In \cref{tab:ablation-label-count}, we report results for increasing $k$ from one-shot to $k=100$. Note that for up to $k=10$, we manually select the samples used for training. For $k > 10$, we randomly add further data. We observe a continuous improvement for greater $k$. Notably, for $k=100$, our pseudo-label generator is almost on par with Panoptic-DeepLab while using \SI{2.9}{\percent} of the annoations (see ResNet-50 backbone in \cref{tab:quantitative-results}).
}

\section{Conclusion}

In this work, we introduced SPINO for few-shot panoptic segmentation by exploiting descriptive image representations from the unsupervised foundation model DINOv2. We demonstrated that SPINO can generate high-qualitative pseudo-labels after being trained on as little as ten annotated images. These pseudo-labels can then be used to train any existing panoptic segmentation method yielding results that are highly competitive to fully supervised learning approaches relying on human annotations. Finally, we extensively evaluated several design choices for the proposed pseudo-label generator and employed our SPINO approach to both public and in-house data.
To facilitate further research, we made our code publicly available. In the future, we will further enhance the instance separation by refining the boundary estimation and employ SPINO in additional domains.

%%%%%%%%%%%%%%%%%%%%%%%%%%%%%%%%%%%%%%%%%%%%%%%%%%%%%%%%%%%%%%%%%%%%%%%%%%%%%%%%

\footnotesize
\bibliographystyle{IEEEtran}
\bibliography{references.bib}

%%%%%%%%%%%%%%%%%%%%%%%%%%%%%%%%%%%%%%%%%%%%%%%%%%%%%%%%%%%%%%%%%%%%%%%%%%%%%%%%

\end{document}